%% file: root.tex
\newif\ifarxiv
\newif\ifralfinal
\ralfinalfalse \arxivtrue

\ifarxiv
\documentclass[letterpaper, 10 pt, conference]{ieeeconf}  
\fi
\ifralfinal
\documentclass[letterpaper, 10 pt, journal, twoside]{IEEEtran}
\fi

\IEEEoverridecommandlockouts

\usepackage{graphics} 
\usepackage{epsfig} 
\usepackage{times} 
\usepackage{amsmath} 
\usepackage{amssymb}  
\usepackage{bbm}
\usepackage{booktabs}
\usepackage{mathtools}
\usepackage{multirow}
\usepackage{units}
\usepackage{bm}
\usepackage{xcolor}
\usepackage{hyperref}
\let\labelindent\relax
\usepackage{enumitem}
\usepackage[ruled,vlined]{algorithm2e}
\usepackage{algorithmicx}
\usepackage{cite} 
\usepackage{subfig}
\usepackage{graphicx}
\usepackage{letltxmacro}
\LetLtxMacro{\originaleqref}{\eqref}
\renewcommand{\eqref}{Eq.~\originaleqref}

\SetCommentSty{mycommfont} 
\usepackage{microtype}
\input{math}

\ifarxiv
\usepackage{fancyhdr}
\fi
\begin{document}
\title{
\ifarxiv\LARGE \bf\fi
Improving Worst Case Visual Localization Coverage via Place-specific Sub-selection in Multi-camera Systems
}

\author{Stephen Hausler\authorrefmark{1}\authorrefmark{2}, Ming Xu\authorrefmark{1}\authorrefmark{2}, Sourav Garg\authorrefmark{2}, Punarjay Chakravarty\authorrefmark{3}, \\ Shubham Shrivastava\authorrefmark{3}, Ankit Vora\authorrefmark{3} and Michael Milford\authorrefmark{2}
\thanks{\authorrefmark{1} these authors contributed equally. \authorrefmark{2} The authors are with the QUT Centre for Robotics, Queensland University of Technology, Brisbane, QLD 4000, Australia (e-mail: \emph{firstname}.\emph{lastname}@qut.edu.au). \authorrefmark{3} The authors are with the Ford Motor Company. This research has been supported by the Ford-QUT Alliance, NVIDIA, the QUT Centre for Robotics and ARC Laureate Fellowship FL210100156.}%
}

\bstctlcite{IEEEexample:BSTcontrol}


\maketitle
\ifarxiv
\thispagestyle{fancy}
\pagestyle{plain}
\fi

\input{tex/0-abstract}
\input{tex/1-introduction}
\input{tex/2-relatedworks}

\input{tex/3-methods}
\input{tex/4-experiments}
\input{tex/5-results}
\input{tex/6-conclusion}


\bibliographystyle{IEEEtran}
\bibliography{references,sg}

\end{document}

%% file: math.tex
\newcommand{\bp}{\mathbf{p}}

\newcommand{\cC}{\mathcal{C}}

\newcommand{\cP}{\mathcal{P}}

\DeclareMathOperator*{\argmin}{arg\,min}

%% file: tex/0-abstract.tex
\begin{abstract}

6-DoF visual localization systems utilize principled approaches rooted in 3D geometry to perform accurate camera pose estimation of images to a map. Current techniques use hierarchical pipelines and learned 2D feature extractors to improve scalability and increase performance. However, despite gains in typical recall$@0.25m$ type metrics, these systems still have limited utility for real-world applications like autonomous vehicles because of their \textit{worst} areas of performance - the locations where they provide insufficient recall at a certain required error tolerance. Here we investigate the utility of using \textit{place specific configurations}, where a map is segmented into a number of places, each with its own configuration for modulating the pose estimation step, in this case selecting a camera within a multi-camera system. On the Ford AV benchmark dataset, we demonstrate substantially improved worst-case localization performance compared to using off-the-shelf pipelines - minimizing the percentage of the dataset which has low recall at a certain error tolerance, as well as improved overall localization performance. Our proposed approach is particularly applicable to the crowdsharing model of autonomous vehicle deployment, where a fleet of AVs are regularly traversing a known route.

\end{abstract}

%% file: tex/1-introduction.tex
\section{Introduction}

\begin{figure}
    \centering
    \includegraphics[width=0.5\textwidth]{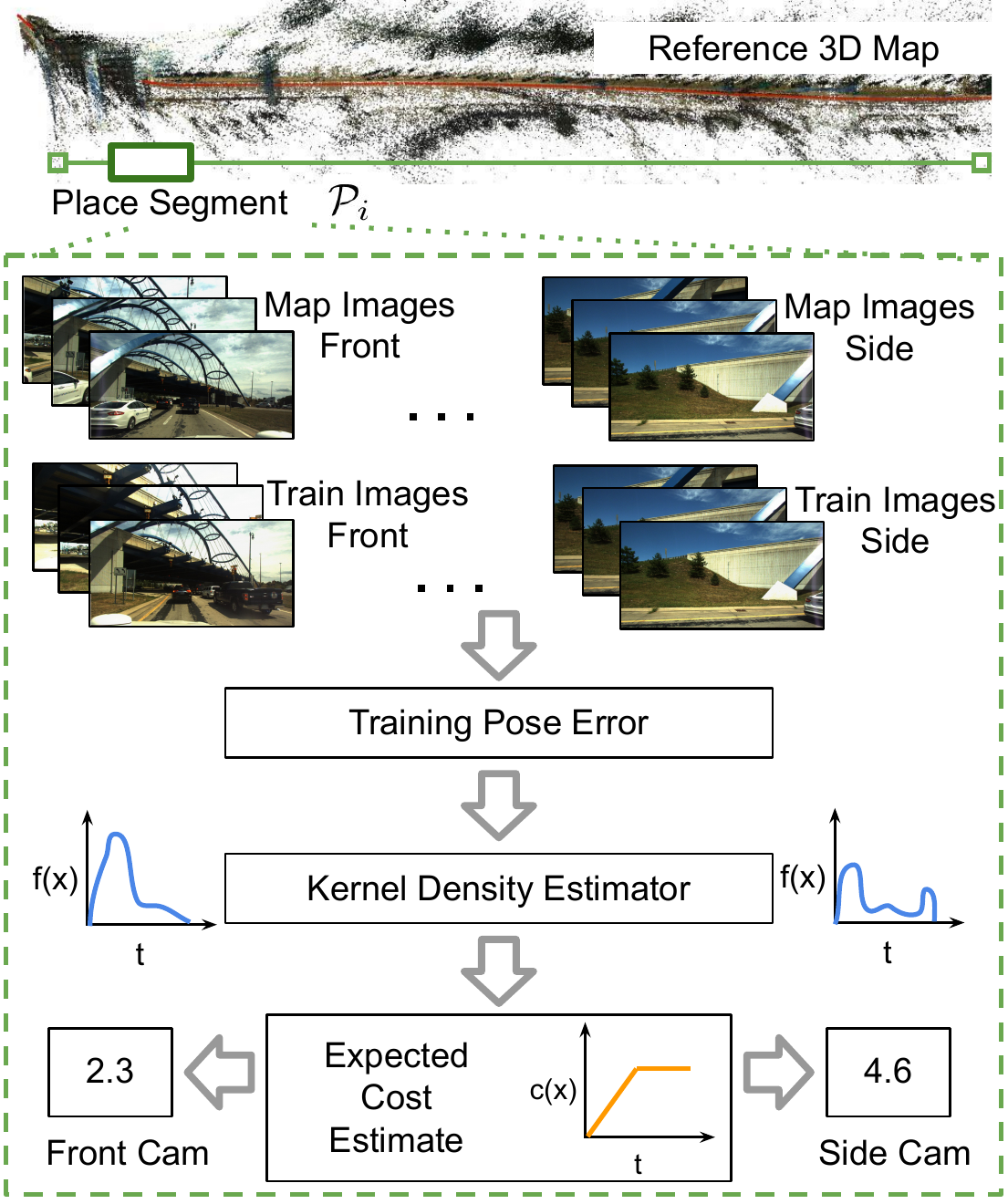}
    \caption{An illustration of the proposed approach for camera selection per place.}
    \label{fig:pipeline_figure}
\end{figure}

\ifralfinal
\else
Autonomous
\fi Vehicle (AV) technology is highly dependent on accurate localization. This localization is typically in the form of cm-precise positioning in a High Definition (HD) map with accurate road network information. Accurate positioning on such a map assists the AV in remaining within the driving lane, follow traffic rules, make safe entrances and exits from freeways and navigate multi-lane intersections and roundabouts. Tyler et al.~\cite{reid2019localization} describe the 95\% accuracy requirement for an SAE Level 3+ AV to be 20 cm (lateral) on freeways and 10 cm (lateral and longitudinal) on urban streets in the USA. 

Currently the most reliable positioning technology used in the industry is a sensor fusion based approach where output from a GNSS+INS integrated system is fused with a Lidar based map matching algorithm using an Extended Kalman Filter~\cite{Wolcott17, vora2020aerial}. In certain geographic regions, a GNSS+INS+RTK corrections (Real Time Kinematics) based system can give sub-5cm accuracies. However, these systems can fail in dense urban environment and in tunnels, meaning they alone are insufficient for reliable autonomous driving. While incorporating a tactical grade INS system~\cite{applanix21} can help, it drives costs up substantially (currently upwards of 70,000 dollars). With recent improvements in visual localization systems ~\cite{Sattler2018,pion2020benchmarking}, cameras offer a potentially cheaper alternative, and are already present in many modern vehicles for Advanced Driver Assistance Systems (ADAS) functions like lane-keeping and parking assistance. Here we build on these recent advances in vision-based localization, presenting research  that further improves their performance on repeated routes through the learning of \textit{place-specific} operating configurations.

State-of-the-art approaches to 6-DoF visual localization assume the availability of a map given as a sparse 3D point cloud reconstruction of an environment. The position of the 3D points within the map are triangulated using images captured across multiple views; this procedure is called structure-from-motion (SfM) and standard methods and software frameworks exist for this process~\cite{schoenberger2016sfm,lindenberger2021pixsfm}. A query image is localized within the map by identifying 2D-3D correspondences and estimating the camera pose using a Perspective-n-Point solver within a RANSAC loop. This process can fail due to a number of issues, including incorrect positioning of 3D points, 2D keypoint localization errors in the query image and incorrect data associations between local-descriptors caused by repetitive environmental structures. A key observation motivating the research presented here is that these issues are often present in varying degrees across multiple cameras situated on a vehicle. Consequently, our objective here is to learn which camera views are most performant in each location during the initial environmental mapping procedure, and then choose camera views accordingly during deployment, as shown in Figure~\ref{fig:pipeline_figure}.

\ifarxiv
\setlength{\topmargin}{-24pt}
\setlength{\headheight}{0pt}
\fi

In this paper we investigate the utility of identifying place specific configurations for camera pose estimation of repeated traverses by utilizing a \textit{training traverse} separate to the mapping traverse. Although our approach requires an additional traverse of the environment, we note this is a reasonable assumption for many autonomous vehicle applications such as ride sharing fleets, where many vehicles traverse the same route each day. In such situations, each query traverse by an autonomous vehicle acts as the training data for all subsequent autonomous vehicles also travelling along the same route.

Our contributions are listed as follows:
\begin{enumerate}
    \item A framework for supervised learning of place-specific, camera pose-estimation configurations when localizing along a regularly traversed route
    \item A specific implementation of this framework where the configuration being learnt and then dynamically chosen is the choice of camera with best localization utility on a multi-camera-equipped vehicle
    \item Demonstration of this system on the benchmark Ford Autonomous Vehicle driving dataset, showing substantially improved worst-case localization performance at all localization error tolerance levels.
    
\end{enumerate}

%% file: tex/2-relatedworks.tex
\section{Related works}

This section reviews relevant research on 6-DoF visual localization.

\subsection{6-DoF Visual Localization}
6-DoF metric localization using vision has gained significant attention recently, particularly aimed at long-term autonomy~\cite{Sattler2018}. Existing works have explored several ways to improve localization performance including utilizing 2D-3D mutual information~\cite{pandey2014toward}, leveraging semantics~\cite{taira2019right, voodarla2021s, garg2020semantics}, multi-task learning~\cite{radwan2018vlocnet++}, including topological information~\cite{badino2012real}, and recent learning-based approaches such as scene coordinate regression~\cite{shotton2013scene, li2020hierarchical} and local feature matching~\cite{Dusmanu2019CVPR,sarlin2020superglue}.

Estimating accurate 6-DoF pose within the practical computational constraints typically requires hierarchical pipelines that involve image retrieval~\cite{pion2020benchmarking}, 2D-2D local feature matching~\cite{DeTone18,sarlin2020superglue} and 2D-3D PnP-based pose estimation using robust optimization~\cite{brachmann2017dsac}. Recent examples of 6-DoF hierarchical localization pipelines include HF-Net~\cite{Sarlin19} and Kapture's localization toolbox~\cite{kapture2020}. While the aforementioned techniques rely on robust local features for wide-baseline image matching, researchers have also explored dense image alignment based techniques for accurate 3D pose estimation. This includes methods like GN-Net~\cite{Stumberg20} and the corresponding benchmark for ``relocalization tracking'' which decouples the image retrieval task from pose estimation. However, these dense methods are sensitive to viewpoint variations and rely on a ``good'' initial coarse estimate. Thus, in this work, we focus on local feature-based 6-DoF localization.

\subsection{Learning-based Methods for Localization}
Several learning-based techniques have been developed recently to improve various components of 6-DoF localization pipelines. This includes learning robust global descriptor methods such as NetVLAD~\cite{Arandjelovic16} and AP-GeM~\cite{revaud2019learning}; learning local keypoints~\cite{Dusmanu2019CVPR, r2d2}, descriptors~\cite{DeTone18,noh2017large}, and matchers~\cite{sarlin2020superglue,rocco2020ncnet} to improve local feature correspondence; joint learning of local and global features~\cite{cao2020unifying}; learning robust optimization~\cite{Stumberg20,sarlin2021back}; and differentiable pose hypothesis selection as a counterpart to vanilla RANSAC~\cite{brachmann2017dsac}. These learning-based techniques require large amounts of training data to learn `general' patterns for the particular sub-task in the whole localization pipeline. Many of the components of these pipelines such as nearest-neighbor search and robust optimization have recently been adapted to suit end-to-end learning~\cite{campbell2020solving,brachmann2017dsac,rocco2020ncnet,sarlin2021back,Stumberg20,gould2021deep}. 
However, they may not necessarily generalize~\cite{zhou2020learn, rocco2018neighbourhood} or may perform inferior to non-learning based methods~\cite{lindenberger2021pixsfm}. In this work, we assume that the target operating environment is seen (revisited) at least once after mapping, which can thus guide and improve localization in subsequent revisits. 

\subsection{Post-processing Prior Maps for Improved Localization}
Since prior map information is always available for (re-)localization tasks, researchers have also investigated various techniques to post-process prior maps using several traverses of the operating environment to improve localization performance. This includes landmark selection based on query-time appearance conditions~\cite{burki2018map, burki2016appearance}, informative reference place selection~\cite{molloy2020intelligent, kovsecka2005global}, relevant ``experience'' selection~\cite{linegar2015work, doan2019scalable}, stable local feature selection~\cite{dymczyk2016will}, and map summarization to improve robustness and efficiency for long-term localization~\cite{muhlfellner2016summary}. In contrast to these methods, we focus on camera selection for a given multi-camera rig, which only requires one additional traverse for training. While past research~\cite{bansal2014understanding} analyzed the influence of camera orientation on localization performance, on-the-fly camera selection has not been explored. Even though generalized camera-based solutions for multi-camera rigs exist for motion estimation~\cite{pless2003using} and pose estimation~\cite{lee2015minimal, geppert2019efficient}, they require highly accurate extrinsic calibration~\cite{heng2015leveraging}. Thus, in this work, we consider an alternative measure for utilizing multiple cameras for precise localization. Since we are investigating multi-camera platforms, multi-sensor literature is also relevant to our approach. Multiple sensors have been used for both coarse localization~\cite{Milford2013a} and fine localization~\cite{Wan2018}, and methods also exist for sensor selection~\cite{SensorSelect2009}.

\subsubsection*{Place- or Location-Specific Methods}
A particular subset of prior-map based techniques comprises those based on place- or location-specific modulations, that is, where post-processing of the map varies for individual places in the map. This has not been a widely-studied area in the context of metric visual localization, and existing methods are either limited to visual place recognition based coarse localization~\cite{keetha2021hierarchical, knopp2010avoiding, carl2012makes, gronat2013learning} or require significant data to train location-specific classifiers~\cite{mcmanus2014scene} and descriptors~\cite{zhang2018learning}. While Linegar et al.~\cite{linegar2016made} use a single mapping traverse to learn place-specific `distinctive landmarks', we leverage the viewpoint-repeatability of AVs traversing repeated routes and propose a method to estimate place-specific camera selection, using a single additional repeat traversal of the operating route.

%% file: tex/3-methods.tex
\section{Methodology}

\subsection{Problem Formulation}

We will now formulate our place-specific camera selection problem. We assume the availability of an SfM reconstruction of the environment which we call the map generated by a vehicle trajectory. First, we partition the vehicle trajectory into a set of $N$ potentially overlapping segments or ``places" $\{\cP_i\}_{i=1}^N$. Each place $\cP_i$ has an associated optimal camera to use for pose estimation $\cC_i^*\in\{1, \dots, N_c\}$, where $N_c$ denotes the number of cameras. 

To determine the optimal cameras to use for each place, we use a training set of images $\{I\}$ from all cameras captured within $\cP_i$ with associated ground truth poses collected from say, LiDAR or RTK GPS. Note, it is a requirement that the method for evaluating ground truth poses is substantially more accurate than the operational requirements for the vision-based system (e.g. LiDAR and RTK GPS). The pose estimation performance for each camera from PnP + RANSAC measured using the training images will be used to select the best camera. Figure~\ref{fig:pipeline_figure} illustrates the steps within the proposed per-place camera selection and localization pipeline.

\subsection{Per-Place Selection of Cameras}

Our proposed methodology for camera selection using the training set revolves around a statistical measure based on expected cost. For a given camera $z$, we have a set of training images $\{I^z\}$ and associated ground truth poses $\{\bp^z\}$. We can apply local feature extraction, matching with mapping images and pose estimation using PnP + RANSAC to yield estimated poses $\{\hat{\bp}_z\}$. We can now compute a measure of pose estimation performance for each training image (we use translation error only in this work) $\{t_{err}^z\}$. This is repeated for all cameras to yield a set of pose error samples for each camera in a given place. 

To select between cameras, we compute an \textit{expected cost} measure for each camera using the pose error samples. This expected cost is defined as
\begin{equation}\label{eq:expcost}
    \mathbb{E}\left[ c(X) \right] = \int_0^\infty c(x)f(x)dx,
\end{equation}
where $x$ denotes pose error, $c(x)$ is a cost function which maps a pose error value to a cost and $f(x)$ is the probability density function over pose errors. In this work, we use monomial cost functions with a fixed ceiling, i.e.
\begin{equation}\label{eq:cost}
    c(x) = \begin{cases}
      x^p & x\leq x_{max} \\
      x_{max}^p & \text{otherwise} 
   \end{cases}
\end{equation}
for $p\geq 0$. Intuitively, higher pose errors map to higher costs with $p>1$ monomials more harshly penalizing high pose errors. We also fix a threshold $x_{max}$ above which the cost no longer increases, to prevent our expected cost from being overly sensitive to outliers which occur when pose estimation fails catastrophically (i.e. PnP finds a spurious pose). We used a quadratic cost (i.e. $p=2$) with $x_{max}=2$ meters under the assumption that a 2 meter error is catastrophic. We ablated KDE parameters (i.e. bandwidth, kernel function) and the degree of the cost functions, i.e. $p\in\{1,2,3\}$ and observed that our final results were not sensitive to these parameters.

We estimate the density over pose error $f(x)$ using kernel density estimators (KDE). KDEs estimate a smooth, empirical density function given a set of samples by placing kernel functions over observations. Concretely, given a set of univariate samples $\{x_i\}_{i=1}^n$, the estimated density is given by
\begin{equation}\label{eq:kde}
    f(x) = \frac{1}{nh} \sum_{i=1}^{n} K \left(\frac{x - x_i}{h}\right)
\end{equation}
where h is the bandwidth parameter and $K$ is a kernel function. We use a Gaussian kernel with a $h=0.1$, noting that our results were not sensitive to variations on $h$. We use the KDE method to estimate $f(x)$ for our pose error samples for each camera. 

Finally, we compute a Monte Carlo estimate of the expected cost \eqref{eq:expcost} by first sampling $N$ samples from $f(x)$ estimated using \eqref{eq:kde} and evaluating the cost using \eqref{eq:cost} for each sample before averaging. We could alternatively use numerical integration to compute \eqref{eq:expcost} directly, but found Monte Carlo to be sufficient for our problem. The best camera is then the camera with the minimum expected cost:
\begin{equation}
    \cC^* = \argmin_{\cC \in\{1, \dots, N_c\}} \{\mathbb{E}_\cC\left[ c(X) \right]\}_{\cC=1}^{N_c},
\end{equation}
where $\mathbb{E}_\cC\left[ c(X) \right]$ is the expected cost for the $\cC^{th}$ camera. Figure~\ref{fig:pipeline_figure} illustrates the components of the methodology.

We note that a key assumption of this method is that the samples of pose estimation performance are statistically independent. We believe this is a reasonable assumption given our definition of a place in this work corresponds to a short trajectory segment (approx. 40m long). This implies the scene is typically similar between views within a place and therefore errors in pose estimation typically arise from random effects induced by robust estimation/image retrieval.

\subsection{Pose Estimation during Query}

During inference time when we wish to estimate the pose of the vehicle using a set of images, one from each camera. We first identify the relevant place within the map using a coarse localization approach, i.e. image retrieval or GPS. Given a place $\cP_q$, we lookup the optimal camera $\cC_q$ obtained from the offline training. We then match local features, generating 2D-3D correspondences and perform pose estimation using PnP + RANSAC.

%% file: tex/4-experiments.tex
\section{Experimental Setup}

We evaluate the utility of our place specific configurations on real-world driving datasets.

\subsection{Datasets}
We use the Ford AV dataset \cite{fordavdata} to evaluate our algorithm. This dataset comprises of data logs collected on a route of approximately 66 km in Michigan, over a span of one year. The 66 km trajectory is split into various parts based on various driving environments. For the testing and validation of our algorithm, we use logs 1,2, 3, 4 and 5 which represent a freeway, airport, urbanized residential, a university region, and a vegetated residential area respectively.

\subsubsection{Camera Images} 
The dataset contains rectified images from 7 cameras in the vehicle. These are Flea3 GigE Point Grey cameras with 12 bit ADC and global shutter. The The two front and the two rear cameras form a 1.3 MP stereo pairs with 3 additional 1.3 MP cameras on each side of the vehicle. There is an additional front camera that produces high resolution 5 MP images at a maximum rate of 6 Hz. All the 1.3 cameras are hardware triggered which ensures time synchronization between them. In our experiments, we use a subset of four cameras, that are \textit{front-left} (FL), \textit{front-right} (FR), \textit{side-left} (SL) and \textit{side-right} (SR).

\subsubsection{Ground Truth} 
We compare our results against the ground truth pose provided by the dataset authors. The ground truth pose is obtained using 6-DoF lidar-based pose-graph SLAM and full bundle adjustment. This method has been previously used by many authors to compare their work against a more accurate source of position information \cite{castorena2017ground, Wolcott17}.

\subsubsection{Geographically and Semantically Disparate Slicing}
The FordAV dataset is comprised of several \textit{logs} which are geographically disparate and consist of images from semantically different types of environment such as multi-lane highways and narrower suburban regions. We used image data from $Log1$, $Log2$, $Log3$, $Log4$ and $Log5$ to represent \textit{Highway}, \textit{Airport}, \textit{Urban}, \textit{University} and \textit{Suburban} categories. Since a complete 3D reconstruction of the whole log is prone to errors~\cite{Sattler2018}, we follow the standard practice of reporting results on different slices for each of the environmental categories. The individual slices are approximately 1 km in length, with individual images spaced $1m$ apart. We also filter and remove any sections of the dataset where the mapping, training and query traverses do not overlap.

\subsubsection{Mapping, Training and Query Setup}
For any given \textit{log}, we considered three traverses of that route for the purpose of mapping (based on sparse 3D reconstruction), training (based on place-specific configuration selection) and querying/testing. All three traverse use the four cameras FL, FR, SL and SR. In order to evaluate robustness against seasonal variations, the mapping and training imagery were captured in August ($2017/08/04$) using vehicle V2 and V3 respectively, whereas the query imagery was captured in October ($2017/10/26$). We also provide an ablation on a 1km subset of Log 1 with query imagery captured in May, which is five years after the map was produced ($2022/05/03$).

\subsection{Implementation Details}

Our camera selection procedure builds upon the standard state-of-the-art visual localization pipeline, provided by open source repositories Kapture~\cite{kapture2020}, Kapture-Localization, COLMAP~\cite{schoenberger2016sfm} and R2D2~\cite{r2d2}. We use R2D2 to extract local keypoints and descriptors from all images, using the default settings. We use COLMAP to generate a SfM reconstruction using these local features. We use the pipeline provided in Kapture-Localization to localize during both training and query, using \emph{Config 1} (as provided in the repository) as the pre-defined RANSAC hyperparameters to pass to COLMAP. At mapping time, we generate a separate SfM reconstruction per camera. Building a joint reconstruction is extremely challenging due to large viewpoint shifts between the front and side cameras. This causes issues in identifying covisible images between cameras as well as local feature matching.

\subsubsection{Configurations of Place Generation}

As detailed in the methodology, we partition the vehicle trajectory into $N$ places $\{\cP_i\}_{i=1}^N$. For all experiments, we define a place as a list of $40$ consecutive images, and each place overlaps with a $3/4$ factor (i.e. each place is $10$ images after the previous). Since we sample images every $1m$, this also means that a place has a width of $40m$ with $10m$ separation between successive places.

\subsection{Evaluation}
We follow the standard evaluation as proposed in~\cite{Sattler2018} for 6-DoF localization, where three different bins of translation and rotation errors are used with a varying degree of localization precision: 0.25 meters and 2\textdegree,  0.50 meters and 5\textdegree, and 5 meters and 10\textdegree. For each of these bins, the percentage of correctly localized query images is reported.

\subsection{Baselines}\label{sec:baselines}
We compare our proposed place specific camera-selection approach on a range of different baseline methods. The first baseline (\emph{Random cam}) is a random camera selection, that is, given a total of $N_c$ cameras we randomly select a camera at every query timestamp. The second baseline (\emph{Static cam}) is a static camera selection, where the best performing camera for a given training slice is selected as the camera to use during the query traverse. The next three baselines consider alternative algorithms for selecting the camera. These three baselines do not require any training data and use a statistic to select the camera on-the-fly by localizing each camera, thus requiring additional processing per camera. These criteria for camera selection are: \emph{Num 3D Points}, based on the highest number of matched 3D map points through local feature matching between the query image and its top retrieved images; \emph{Inlier Count}, based on the largest inlier count after pose estimation using PnP-RANSAC on each camera individually; and \emph{Inlier Ratio}, which is the same as the previous but uses inlier ratio instead. Finally, \emph{Multi-cam GPnP} is used as a multi-camera joint pose estimation method based on GPnP implementation from Poselib.\footnote{https://github.com/vlarsson/PoseLib}

%% file: tex/5-results.tex
\section{Results}\label{sec:results}

\subsection{Quantitative Performance Comparisons}

\begin{figure*}[t!]
    \centering
    \includegraphics[width=0.8\textwidth]{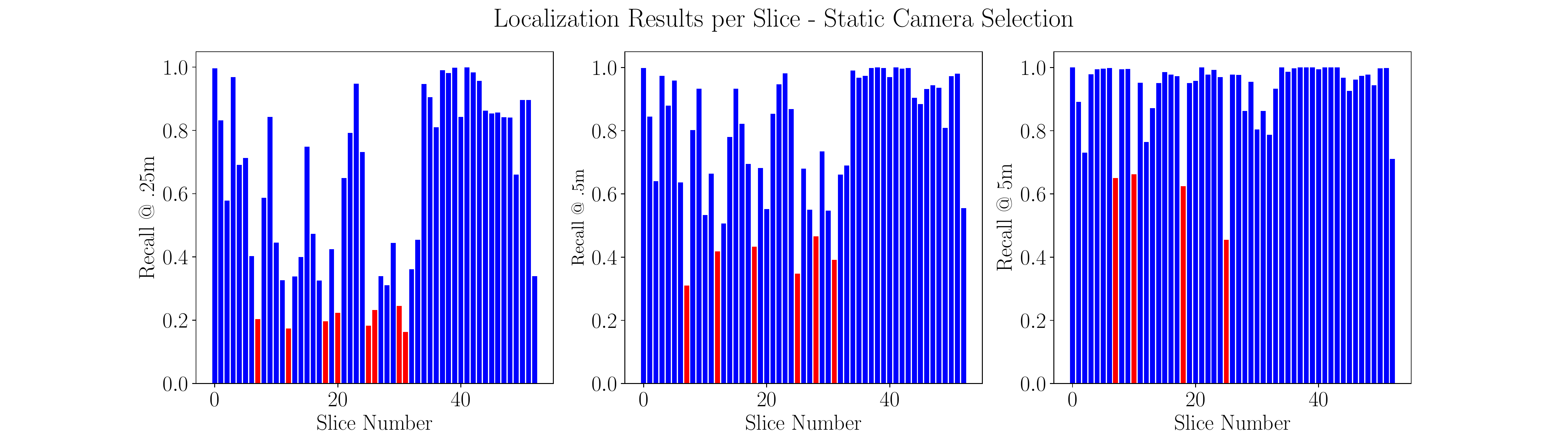}
    \caption{Localization performance with a static best camera at the three tolerances $.25m$, $.5m$ and $5m$, for every $1km$ slice across all five logs of the Ford AV dataset. We denote AV system failures as red bars, where a failure is a recall rate less than $30\%$ at $0.25m$, less than $50\%$ at $0.5m$ and less than $70\%$ at $5m$. This figure should be compared with Figure \ref{fig:allslicescalib}.}
    \label{fig:allslicesstatic}
    \vspace*{-0.2cm}
\end{figure*}

\begin{figure*}[t!]
    \centering
    \includegraphics[width=0.8\textwidth]{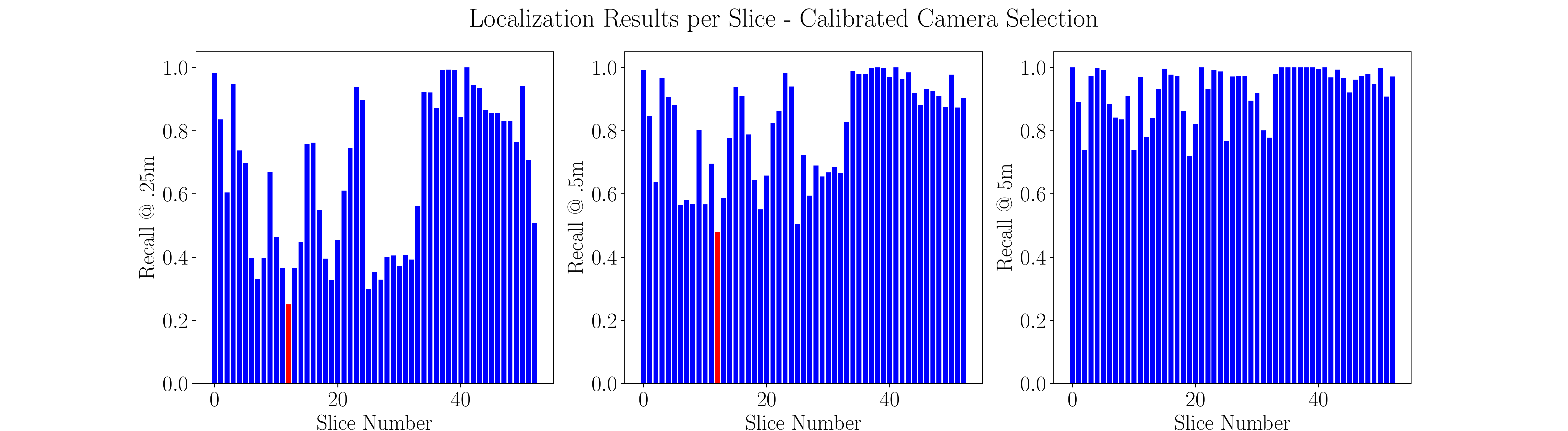}
    \caption{Localization performance with the dynamic camera selection system at the three tolerances $.25m$, $.5m$ and $5m$, for every $1km$ slice across all five logs of the Ford AV dataset. We denote AV system failures as red bars, where a failure is a recall rate less than $30\%$ at $0.25m$, less than $50\%$ at $0.5m$ and less than $70\%$ at $5m$. Compared to the best overall static camera choice (Figure \ref{fig:allslicesstatic}), the number of below acceptable recall rate failures is substantially reduced.}
    \label{fig:allslicescalib}
    \vspace*{-0.2cm}
\end{figure*}

In this section we analyse the effect of selecting different cameras at different places in a deployment environment, where the selection rule is based on a training traverse and not on the test traverse itself. Our training-based camera selection policy will be referred to as `dynamic camera' in the remaining text. Results are analysed at both the \emph{per log} and \emph{per slice} levels; a log is a contiguous set of slices with common environmental characteristics, while a slice is a $1$ km subset of the full navigation experiment. We analyse $53$ slices in total, with the localization performance provided in Figures~\ref{fig:allslicesstatic} and \ref{fig:allslicescalib}. Figure~\ref{fig:allslicesstatic} shows the performance with the best static camera. The best static camera is determined as the overall best performing individual camera, as determined on the training traverse on a per-slice basis and referred to as `static camera' from here on. Figure~\ref{fig:allslicescalib} details the localization capabilities with our dynamic camera selection algorithm, which selects the best camera per place $\cP_i$. We consider a slice to have `failed' to adequately localize if the recall is less than a pre-defined threshold for a given localization tolerance. Our thresholds are $30\%$ for $.25m$, $50\%$ for $0.5m$ and $70\%$ for $5m$, based on operational requirements in the application domain. The thresholds increase with error tolerance levels since larger localization errors are more detrimental to AV performance~\cite{Rehrl2021}. In Figures~\ref{fig:allslicesstatic} and \ref{fig:allslicescalib}, failed slices are marked in red. In summary, the best static camera fails to adequately localize for $8$ slices out of $53$ ($\mathbf{15}\%$) at a tolerance of $0.25m$, for $6$ slices out of $53$ ($\mathbf{11}\%$) at a tolerance of $0.5m$, and $4$ slices of out $53$ ($\mathbf{7.5}\%$) at a tolerance of $5m$. Our dynamic calibration system, in comparison, fails to localize for $1$ slice out of $53$ ($\mathbf{1.9}\%$) at a tolerance of $0.25m$, for $1$ slice out of $53$ ($\mathbf{1.9}\%$) at a tolerance of $0.5m$, and no slices fail at a tolerance of $5m$. Dynamic camera selection substantially reduces the worst-case localization performance.

We also include a summary of the localization performance on a per log basis, as provided in Table~\ref{tab:summaryPerLog}. We compare the dynamic camera against several baselines, as detailed in Section~\ref{sec:baselines}. Our summary results show that the dynamic camera has the most precise localization, with the highest recall at the $0.25m$ threshold, on average across the five logs. The highway environment (Log1) has the greatest improvement after training - one theory is that the lack of distinctive visual features in a multi-lane highway means that alternative cameras are more often required in order to find sufficiently accurate 2D and 3D local features for localization. 

We find that the training-free camera selection baselines (\textit{Num 3D Points}, \textit{Inlier Count}, \textit{Inlier Ratio}) achieve superior performance on average compared to both a randomly selected camera and a static camera, which indicates that these new selection rules are able to select the best camera with some measure of success. We note that these baselines have reduced localization performance at the $0.25m$ tolerance compared to our proposed camera selection algorithm, which uses an additional training traverse to select the best camera per place. We hypothesise that having access to training data provides more accurate information for camera selection. We notice that these new baselines perform better at looser tolerances, we hypothesise this could be because query-time camera selection is avoiding cameras that are occluded by dynamic objects. Future work will investigate a hybrid strategy, which combines both training-based and query-time camera selection. However, these baselines are not as computationally efficient as the dynamic camera, since they require local feature extraction, feature matching, and pose estimation \textit{per camera} at query time, whereas the dynamic camera only needs to execute the localization pipeline for the selected camera at the current place $\cP_i$.

We observe that the multi-camera localization baseline performs suboptimally at tight tolerances, which we hypothesise is due to errors incurred by poorly performing cameras. For example, if the side camera is observing a large number of occlusions and is performing poorly during localization by itself, then the GPnP result will potentially be skewed if the final set of inlier points include inliers from this camera. This provides further evidence that some form of camera sub-selection is necessary on multi-camera rigs.

\input{tables/5a_results_summary_postrevisions}\label{table:sum}

\subsection{Ablation Studies}

To evaluate the generalization capabilities of our approach, we ran an ablation study considering whether the place-specific (dynamic) camera selection, based on a training traverse, is still applicable after a significant amount of time has passed. These new results are shown in Table~\ref{tab:query5year}. This new query traverse is challenging, due to the five year gap to the mapping and training traverses, plus the traverse is in rainy conditions. We observe a substantial reduction in localization performance across all techniques. The dynamic camera selection localizes similarly to the static camera, but the benefits of our approach disappear when such a large time span exists between the training and query traverse. However, we notice that alternate camera selection rules (\textit{Num 3D Points}, \textit{Inlier Count}, \textit{Inlier Ratio}) are able to improve the localization accuracy compared to a static camera. This is because these alternatives make camera selections just using the query traverse, albeit at a higher computational cost compared to our training-based dynamic camera selection.

\input{tables/5c_results_newseason}\label{table:new}

\subsection{Qualitative Examples}

\begin{figure}[t]
    \centering
    \includegraphics[width=0.95\linewidth]{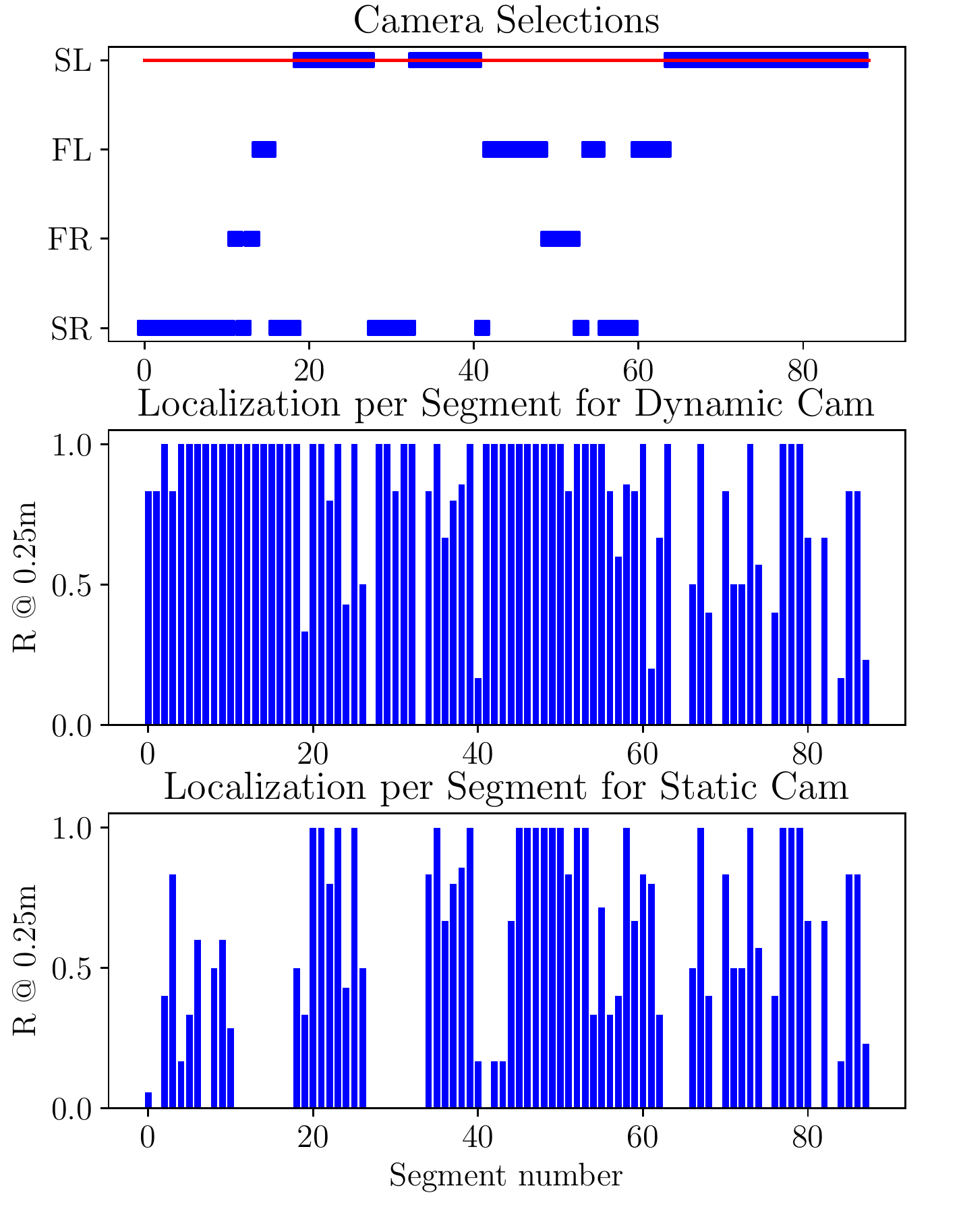}
    \caption{\emph{Top:} for each segment (place) in a slice, we show the camera selection that was chosen, as indicated by a blue square. The red line denotes the static camera for this slice. \emph{Mid:} We plot the recall rate at $0.25m$, based on the individual pose errors (at query time) contained within each segment using the training-based system (higher recall is better). \emph{Bot:} We compare against the localization performance using a static camera, in this case, camera SL.}
    \label{fig:qual}
    \vspace*{-0.2cm}
\end{figure}

In this section we provide a qualitative analysis of the dynamic camera selection algorithm. In Figure~\ref{fig:qual}, we plot the camera selections for one slice of the dataset. Each overlapping $40m$ segment (place) has its own selected camera, marked as a blue square. We also mark the static camera for this slice using a red line. We show the localization performance (during the query traverse) on a per segment basis, to demonstrate the difference in performance as the chosen camera is varied. The camera selection algorithm is able to prevent the localization failures that the static camera suffers from between segments $11$ and $17$ and segments $28$ and $32$ inclusive.

\begin{figure}[t]  
    \centering
    \includegraphics[width=1.0\linewidth, trim=1.7cm 1cm 1.7cm 1cm,clip]{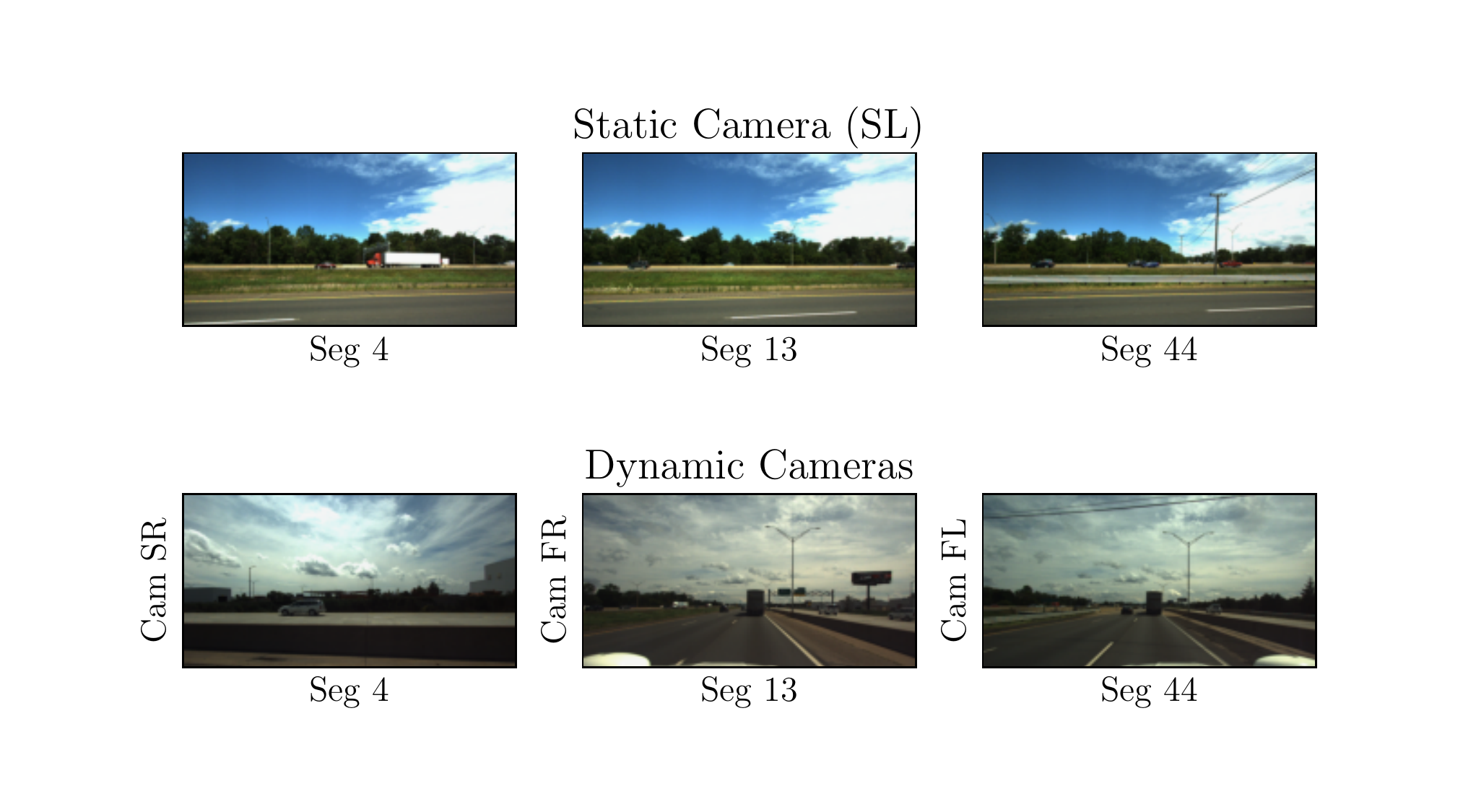}
    \caption{Sample images from the Ford AV dataset, from both the static camera and the chosen dynamic camera for segments 4, 13 and 44. We use the same slice as used in Figure~\ref{fig:qual}.}
    \label{fig:images}
    \vspace*{-0.2cm}
\end{figure}

In Figure \ref{fig:images}, we show image pair examples showing what the vehicle is seeing with different cameras, at locations where the dynamic selection procedure chooses a different camera to what is the best overall static camera for the slice. For direct comparison to the recall performance, we use the same slice as used in Figure~\ref{fig:qual}. In all pairs, the training process has determined that the best overall static camera is SL. In the first pair, the dynamically selected camera is SR. In the second pair, the dynamically selected camera is FL and in the final pair, the dynamically selected camera is FR. We observe that the dynamic camera tends to contain more semantically meaningful features in the images (such as buildings and road signs) compared to the static camera. In these three cases, using the dynamically chosen camera results in higher recall at $0.25m$ compared to the static camera (see Figure~\ref{fig:qual}), for these segments.

%% file: tables/5a_results_summary_postrevisions.tex
\begin{table*}
  \footnotesize
  \setlength\tabcolsep{0.05cm}
      \caption{Summary Results per Log}
    \centering
    \resizebox{\textwidth}{!}{%
    \begin{tabular}{l | ccc | ccc | ccc | ccc | ccc}
    \toprule
    \multirow{2}{*}{Method/Dataset} &
    
    \multicolumn{3}{c}{Log1 - Highway} &
    \multicolumn{3}{c}{Log2 - Airport} &
    \multicolumn{3}{c}{Log3 - Urban} &
    
    \multicolumn{3}{c}{Log4 - University} & 
    \multicolumn{3}{c}{Log5 - Suburban} \\
    \cmidrule{2-16}
    &
    .25m/2\textdegree & .5m/5\textdegree & 5m/10\textdegree &
    .25m/2\textdegree & .5m/5\textdegree & 5m/10\textdegree &
    .25m/2\textdegree & .5m/5\textdegree & 5m/10\textdegree &
    .25m/2\textdegree & .5m/5\textdegree & 5m/10\textdegree &
    .25m/2\textdegree & .5m/5\textdegree & 5m/10\textdegree \\
    \midrule
    Random cam & 
    48.5 & 71.2 & 92.0 &
    51.3 & 71.2 & 91.0 &
    90.6 & 98.0 & 99.3 &
    72.9 & 80.6 & 85.4 &
    76.9 & 89.6 & 95.6 \\
    
    Static cam & 
    52.6 & 71.6 & 90.1 &
    55.4 & 73.7 & 90.4 &
    94.5 & \textbf{99.2} & 99.8 &
    85.0 & 92.0 & 96.0 &
    72.1 & 84.5 & 92.0 \\
    
    Num 3D pts cam select &
    52.7 & \textbf{80.3} & 99.5 &
    57.3 & 77.9 & 96.9 &
    92.7 & 99.1 & 99.7 &
    85.2 & 92.5 & 97.0 &
    \textbf{75.4} & 91.9 & 99.9 \\
    
    Inlier count cam select &
    53.1 & 80.0 & 99.3 &
    58.4 & \textbf{79.9} & 96.9 &
    92.5 & 98.9 & \textbf{99.9} &
    84.8 & 92.5 & 97.4 &
    75.3 & 91.9 & 99.9 \\
    
    Inlier ratio cam select &
    53.0 & 79.4 & 99.5 &
    58.4 & \textbf{79.9} & 96.9 &
    92.8 & 99.0 & 99.7 &
    \textbf{85.4} & \textbf{93.2} & 97.0 &
    75.2 & \textbf{92.6} & 99.9 \\
    
    Multi-cam GPNP &
    28.3 & 46.8 & \textbf{99.9} &
    30.5 & 46.6 & \textbf{97.5} &
    45.3 & 77.3 & \textbf{99.9} &
    55.0 & 77.8 & \textbf{98.7} &
    28.6 & 56.6 & \textbf{100} \\
    
    Dynamic cam (Ours) & 
    \textbf{58.7} & 78.0 & 93.0 &
    \textbf{58.7} & 74.8 & 89.5 &
    \textbf{94.7} & 98.6 & 99.4 &
    84.6 & 91.3 & 95.9 &
    75.1 & 91.6 & 96.1 \\
    
    \bottomrule

    \end{tabular}
    }
    \label{tab:summaryPerLog}
\end{table*}

%% file: tables/5c_results_newseason.tex
\begin{table}
  \footnotesize
  \setlength\tabcolsep{0.05cm}
      \caption{Results for Query 2022-05-03}
    \centering
    \begin{tabular}{l | ccc}
    \toprule
    \multirow{2}{*}{Method/Dataset} &
    
    \multicolumn{3}{c}{Log1 - 1km Slice} \\
    \cmidrule{2-4}
    &
    .25m/2\textdegree & .5m/5\textdegree & 5m/10\textdegree \\
    \midrule
    Random cam & 
    1.1 & 17.4 & 92.6 \\
    
    Static cam & 
    0.7 & 11.8 & 96.0 \\
    
    Num 3D pts cam select &
    1.8 & 20.1 & 96.8 \\
    
    Inlier count cam select &
    1.5 & 18.8 & 95.4 \\
    
    Inlier ratio cam select &
    1.5 & 16.1 & 94.5 \\
    
    Multi-cam GPNP &
    0.9 & 12.2 & 100 \\
    
    Dynamic cam (Ours) & 
    0.5 & 11.2 & 95.3 \\
    
    \bottomrule

    \end{tabular}
    \label{tab:query5year}
\end{table}

%% file: tex/6-conclusion.tex
\section{Discussion and Conclusion}

In this work we investigated a novel concept that learns location-specific operating configurations in order to maximize visual localization system performance, in this case which camera to use at a specific location. Our system chooses from multiple cameras on the AV, but requires only the computational overhead for processing a standard single camera system. We found that such a dynamic camera selection process significantly decreases the occurrence of periods of poor localization performance where recall drops below a certain threshold, compared to using a single static camera that was determined during training to be the best on average. For example, at a 0.5m error tolerance, the length of road where recall drops below 50\% decreases from $6km$ to $1km$.  

There are a number of avenues for future work. Firstly, the training traverse concept, where a second traverse after the mapping traverse is used prior to deployment, can be expanded to enhance other configurations beyond just camera selection. These configurations could include RANSAC hyperparameters, subsets of local features, image region selection, or even the process of using a semantic segmentation network and checking for semantic label consistency of 3D points. Future work could also expand the camera selection approach (or any other system configuration) to continually learn (and update) the optimal configuration to use in each locality based on ongoing, repeated traverses of the environment, leveraging either repeated traverses by a single vehicle such as a `last mile' autonomous shuttle or fleets of rideshare AVs traversing the same route. Finally, the system could be enhanced by combining both training-time and query-time camera selection, selecting the best camera from a consensus of selection criteria.